\documentclass{article}

\usepackage{microtype}
\usepackage{graphicx}
\usepackage{subfigure}
\usepackage{booktabs} % for professional tables

\usepackage{macros}

\usepackage{hyperref}

\usepackage[accepted]{icml2024}

\usepackage{amsmath}
\usepackage{amssymb}
\usepackage{mathtools}
\usepackage{amsthm}

\usepackage[capitalize,noabbrev]{cleveref}

\theoremstyle{plain}

\theoremstyle{definition}

\theoremstyle{remark}

\usepackage[textsize=tiny]{todonotes}

\icmltitlerunning{Spectrum-Informed Multistage Neural Network: Multiscale Function Approximator of Machine Precision, ICML 2024}

\begin{document}

\twocolumn[
\icmltitle{Spectrum-Informed Multistage Neural Network: Multiscale Function Approximator of Machine Precision}

% It is OKAY to include author information, even for blind
% submissions: the style file will automatically remove it for you
% unless you've provided the [accepted] option to the icml2024
% package.

% List of affiliations: The first argument should be a (short)
% identifier you will use later to specify author affiliations
% Academic affiliations should list Department, University, City, Region, Country
% Industry affiliations should list Company, City, Region, Country

% You can specify symbols, otherwise they are numbered in order.
% Ideally, you should not use this facility. Affiliations will be numbered
% in order of appearance and this is the preferred way.
\icmlsetsymbol{equal}{*}

\begin{icmlauthorlist}
\icmlauthor{Jakin Ng}{stanford,mit}
\icmlauthor{Yongji Wang}{stanford,nyu}
\icmlauthor{Ching-Yao Lai}{stanford}
\end{icmlauthorlist}

\icmlaffiliation{mit}{Massachusetts Institute of Technology, Cambridge, MA.}
\icmlaffiliation{stanford}{Stanford University, Palo Alto, CA.}
\icmlaffiliation{nyu}{New York University, New York, NY.}

\icmlcorrespondingauthor{Ching-Yao Lai}{cyaolai@stanford.edu}
\icmlcorrespondingauthor{Yongji Wang}{yw8211@nyu.edu}
\icmlcorrespondingauthor{Jakin Ng}{jakinng@mit.edu}

% You may provide any keywords that you
% find helpful for describing your paper; these are used to populate
% the "keywords" metadata in the PDF but will not be shown in the document
\icmlkeywords{Machine Learning, ICML}

\vskip 0.3in
]

% this must go after the closing bracket ] following \twocolumn[ ...

% This command actually creates the footnote in the first column
% listing the affiliations and the copyright notice.
% The command takes one argument, which is text to display at the start of the footnote.
% The \icmlEqualContribution command is standard text for equal contribution.
% Remove it (just {}) if you do not need this facility.

\printAffiliationsAndNotice{}  % leave blank if no need to mention equal contribution
% 

% \printAffiliationsAndNotice{\icmlEqualContribution} % otherwise use the standard text.

\begin{abstract}
Deep learning frameworks have become powerful tools for approaching scientific problems such as turbulent flow, which has wide-ranging applications. In practice, however, existing scientific machine learning approaches have difficulty fitting complex, multi-scale dynamical systems to very high precision, as required in scientific contexts. We propose using the novel multistage neural network approach with a spectrum-informed initialization to learn the residue from the previous stage, utilizing the spectral biases associated with neural networks to capture high frequency features in the residue, and successfully tackle the spectral bias of neural networks. This approach allows the neural network to fit target functions to double floating-point machine precision $O(10^{-16})$. 
\end{abstract}

\section{Introduction}
\label{intro}

\subsection{Precision machine learning}
Typical machine learning applications, such as computer vision or natural language processing, do not necessarily require neural networks to fit data to extremely high precision. For instance, the training loss function may simply be a proxy for the true metric, and moreover noise present in the training data may cause models with very low training loss to overfit \cite{michaud2023precision}. Recently, deep learning techniques are being increasingly developed for scientific purposes where high precision is desirable or required \cite{wang2023asymptotic,michaud2023precision,wang2024multi,muller2023achieving,jnini2024gauss}. For instance, neural networks used as interpolators for equation discovery, which infers an exact equation or formula based on data, the correctness of the learned equation requires very high accuracy across multiple scales \cite{udrescu2020ai,mojgani2024interpretable}. Another application requiring high precision is physics-informed neural networks (PINNs), which behave as numerical solvers for partial differential equations (PDEs), where high accuracy is an intrinsic requirement. When solving a well-posed PDE problem, there exists a theoretical global minimum for the PINN training where the training loss should converge to zero \cite{raissi2019physics}. 

In this study, we focus on studying the regression problem as a preliminary example to demonstrate the challenge of training neural networks to approach a target function with high precision. The regression problem is given as: given a dataset $\{(\mathbf{x_i}, u_i = u(\mathbf{x_i}))\}$ sampled from a continuous target function $u$, the neural network $\CN_\theta$, with parameters $\theta$, is trained to fit the data points to approximate the function $u$. We assume that the number of data points is sufficient, and that each of them has zero noise, which guarantees that the data set contains sufficient authentic information of the target functions. Although the universal approximation theorem is a theoretical guarantee that neural networks of large enough size can approximate any function arbitrarily well \cite{hornik1989multilayer}, in practice, the training loss is easily trapped in local minima and eventually plateaus after a certain number of iterations. Many advanced techniques exist to expedite training \cite{sitzmann2020implicit,michaud2023precision,liu2020multi}, but are unable to consistently reduce the training error down to the required precision, which is one of the significant challenges of using deep learning methods for scientific objectives. In contrast, classical numerical methods can consistently reduce error, for instance by increasing the mesh resolution.

\subsection{Spectrum-informed initialization for regression} 

\begin{figure*}[t!]
    \centering
\includegraphics[width=\linewidth]{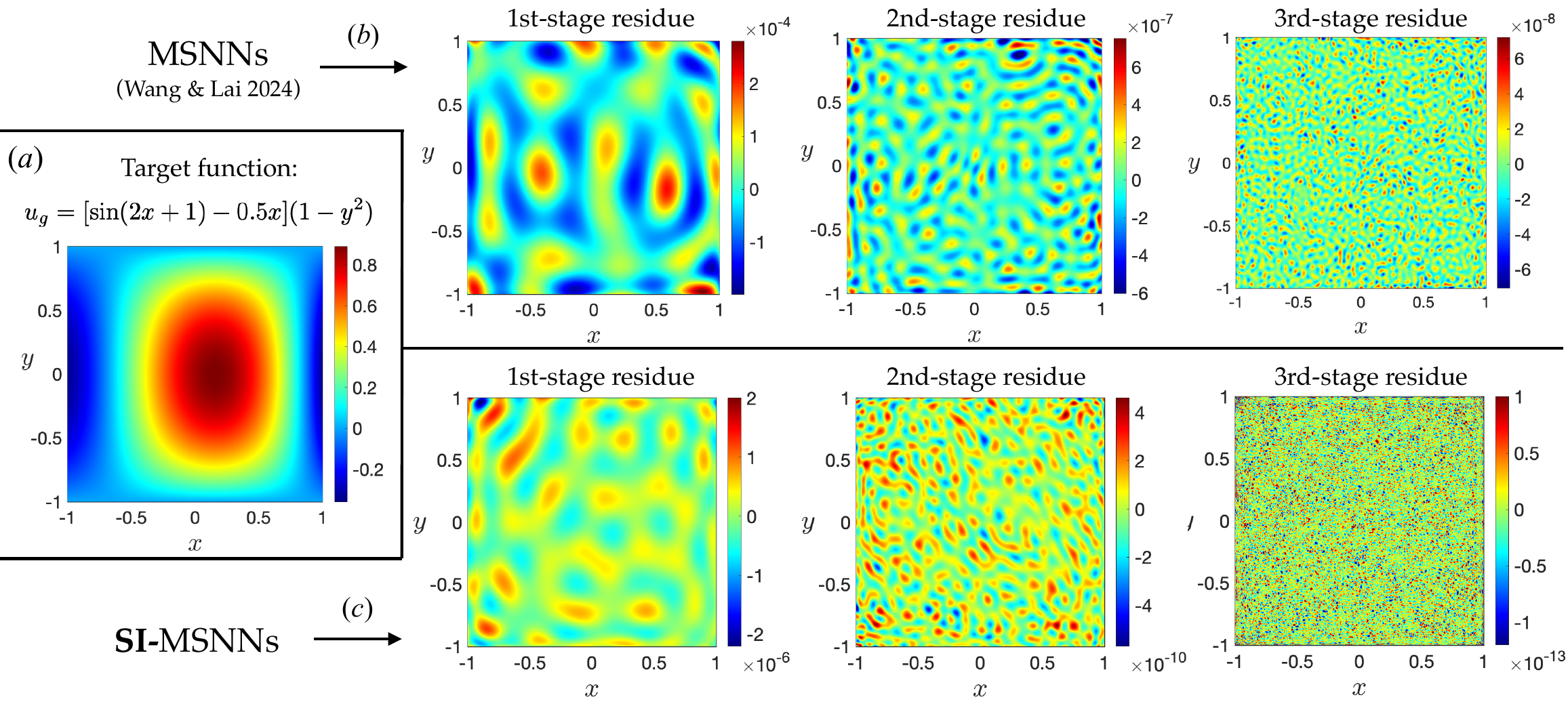}
    \caption{($a$) Target function $u_g(x, y)$ for a 2-D regression problem. ($b$) Errors of neural networks with the target function $u_g$ after different stages of training using the original multi-stage neural networks (MSNNs) where the scaling factor was used to mitigate the spectral biases of the network training. The 3rd-stage residue reaches $O(10^{-8})$. ($c$) Errors after different stages of training using spectrum-informed initialization for both weights and biases of the network, which reaches $O(10^{-13})$ by the 3rd stage of training.}
    \label{fig:msnn-comparison}
\end{figure*}

To resolve the precision limit of PINNs, \citet{wang2024multi} proposed the multistage training scheme which divides neural network training into different stages. For each stage of training, a new neural network was introduced and optimized to learn the residues from previous stages, which largely increases the convergence rate from linear decay to approximately exponential decay. The combined neural networks after several stages of training can approximate the target function up to double floating-point machine precision $O(10^{-16})$ for one-dimensional problems. However, this fails to hold for two- or higher-dimensional problems (Figure \ref{fig:msnn-comparison}$a$). This indicates that more advanced techniques are required to enhance the multistage neural networks (MSNNs) to ensure high-precision approximation for higher-dimensional problems.

The training of neural networks is known to suffer from spectral bias, which is a phenomenon in which neural networks tend to fit low-frequency features of the target function, and may fail to capture high-frequency information \cite{rahaman2019spectral,xu2022overview}. Spectral bias is particularly problematic when using neural networks as function approximators for multiscale problems, such as turbulence \cite{mojgani2024interpretable,rybchuk2023ensemble,chattopadhyay2023long,lai2024machine}. Various frequency domain approaches have been proposed to mitigate this problem, including the scale factor approach \cite{cai2019phasednn,liu2020multi,jin2024fourier,li2023deep}, which were applied in the multistage neural networks \cite{wang2024multi} by multiplying a large scale factor to the weight between the input and first hidden layer. The optimal value of the scale factor was found to be $\kappa = \pi f_d / \sqrt{V_{ar}}$, where $f_d$ is the domain frequency of the target function and $V_{ar}$ is the variance of first layer weights. Although this setting works effectively for fitting one-dimensional functions, it remains limited when approximating high-frequency function in higher dimensions. Here, to resolve the issue, we provide an advanced method that can further alleviate spectral bias and enhance the multistage neural network training, by initializing the neural network weights based on more straightforward spectral information from the target function. In more details, this spectrum-informed initialization uses the discrete Fourier transform of the target function or the residues to inform the initialization of the neural network, ensuring fast convergence across the spectral modes present. In doing so, the spectrum-informed multistage neural network (SI-MSNN) is able to approximate the target function down to $O(10^{-16})$. Moreover, the regression provided by the neural network is accurate across the Fourier spectrum, even for challenging multi-scale target functions.

In Section \ref{preliminaries}, we provide an overview of Fourier feature embeddings, the discrete Fourier transform, and neural tangent kernel theory. In Section \ref{methods}, we introduce the spectrum-informed initialization of network weights, and its usage for multi-stage neural networks (MSNNs). In Section \ref{experiments}, we provide experimental results demonstrating the advantage of SI-MSNNs over the original MSNNs on error reduction and its ability to reach machine precision for complex regression problems. Lastly, in Section \ref{discussions}, we provide further discussions and future work on the spectrum-informed initialization for multistage neural networks.

\section{Preliminaries}\label{preliminaries}
\subsection{Discrete Fourier transform}

The Fourier transform decomposes a function into the various frequencies and corresponding amplitudes that are present, writing the function in terms of a complex exponential or sinusoidal/cosinusoidal basis \cite{sneddon1995fourier}. The discrete Fourier transform (DFT) of a function with values provided at $N$ equally spaced points can be computed using the fast Fourier transform (FFT) algorithm in $O(N \log N)$. The one-dimensional DFT of a function $u$, specified at $N$ points $x_n$, is given by \begin{equation}\tilde u(k) = \frac{1}{N}\sum_{n = 0}^{N - 1} u(x_n)\exp(-i k x_n),\end{equation} for which the inverse Fourier transform is \begin{equation}\tilde u(k) = \sum_{n = 0}^{N - 1} u(x_n)\exp(i k x_n),\end{equation} where $k$ is the wavenumber. Analogously, the two-dimensional DFT of a function specified on a grid with $N_x$ values $x_n$ and $N_y$ values $y_m$ is given by \begin{equation}\tilde u(\mathbf{k}) = \sum_{n = 0}^{N_x}\sum_{m = 0}^{N_y}\exp(-i(k_x x_n + k_y y_m)),\end{equation} where $\mathbf{k} = (k_x, k_y)$ is the wavenumber vector. 

\subsection{Fourier feature networks}
Spectral bias, which is also referred to as the frequency principle, is a well-known phenomenon of standard multi-layer perceptrons (MLPs), which tend to exhibit a bias towards fitting low-frequency information in target functions, and learn high-frequency information more slowly, or fail to converge on high frequency modes \cite{rahaman2019spectral,xu2022overview}. To allow MLPs to learn high-frequencies, Fourier feature networks first pass input points $x$ through a Fourier feature mapping $\gamma(x)$, before passing the result through the MLP \cite{rahimi2007random,tancik2020fourier}. The Fourier feature mapping $\gamma(x)$ can be alternatively considered as the first layer of the MLP, as a composition of a nonlinear, periodic activation function and a linear function with added bias weights. 

\subsubsection{Random Fourier features} Random Fourier features are a non-trainable input mapping
\begin{equation}\gamma(x) = \cos(2\pi B x + b)\end{equation} where $B$ and $b$ are fixed random variables, and the cosinusoidal function is applied component-wise \cite{rahimi2007random}.

\subsubsection{Fourier feature mapping} Fourier feature mapping extends this technique, using the positional encoding 
\begin{equation}\gamma(x) = [A\cos(2\pi B x), A\sin(2\pi Bx)]^T,\end{equation} where $B \in \RR^{m \by d}$ are the weights for the feature mapping, which can be either fixed or trainable. Typically, these weights are initialized as $B \sim \CN(0, \sigma^2)$, where $\sigma$ is a hyperparameter that determines the span of the initialized weights and can be set to 1 as a default, or determined based on the problem \cite{tancik2020fourier}. 

\subsubsection{Cosinusoidal activation} Since cosine and sine differ only by a phase, which can be recovered using the trainable phase parameter $b$, it is comparable to use the cosinusoidal mapping \begin{equation}\gamma(x) = [A\cos(2\pi B x + b)]^T,\end{equation} where the weights $B$ are initialized as in the previous section, and the biases $b$ can be initialized to zero or following a normal distribution. Equivalently, the cosinusoidal feature mapping can be seen as a modified first layer of an MLP, where the activation function is $\sigma(x) = A\cos(2\pi x)$ rather than the typical choices $\sigma(x) = \tanh(x)$ or $\sigma(x) = \operatorname{ReLU}(x).$  

\subsection{Neural tangent kernel}
The neural tangent kernel (NTK) theory describes the evolution of neural networks by gradient descent during training. In the limit of an infinite-width neural network, the NTK can be used to approximate the result of training a neural network as the learning rate approaches zero. In fact, the neural network converges in the eigenvectors of the NTK at an exponential rate with respect to the corresponding eigenvalue \cite{jacot2018neural,tancik2020fourier}. For a conventional MLP, the spectral bias can be viewed as an eigenvector bias, as the eigenvalues decay rapidly \cite{wang2021eigenvector}. When using a Fourier feature mapping, the principal eigenvectors of the NTK correspond to the embedded frequencies from $B$, and thus embedding the frequencies of the target function $u$ into the neural network using Fourier feature mapping allows rapid convergence in those directions \cite{tancik2020fourier}.

\section{Methods}\label{methods}

\subsection{Problem setup}

We consider the supervised learning regression problem, where $N_d$ data points $\{(\mathbf{x_i}, u(\mathbf{x_i}))\}$ are provided, and the task is to fit a neural network to the target function $u$. In the precision machine learning setting, we assume that the data is exact, in order to test the capacity of neural networks as universal function approximators. 

A neural network with $L$ hidden layers, given a feature embedding $\gamma(x)$, is a mapping $\CN: \RR^d \to \RR$ which acts as $\CN(x; \theta) = f_L \circ \sigma_L \circ \cdots \cdots \circ \sigma_1 \circ f_1 \circ \gamma(x)$, where $f_k(x) = W_kx + b_k$ is a linear function and $\sigma_k(x)$ is a nonlinear activation function, such as $\tanh$ or $\operatorname{reLU}$.

\subsection{Multi-stage neural networks (MSNN)}

As mentioned above, multi-stage neural networks \cite{wang2024multi} are a novel approach dividing the training into stages, maintaining a high rate of convergence and allowing the regression for one-dimensional problems to reach machine precision \cite{wang2024multi}. 

 Given a target function $u$, a typical neural network $u_0(x)$ with Xavier weight initialization is trained to approximate $u / \epsilon_0$, with the input coordinates $\mathbf{x}$ normalized to within $[-1, 1]$. The normalizing factor $\epsilon_0$ is taken to be the root mean square value $\epsilon_0 = \sqrt{\frac{1}{N_d}\sum_{i = 0}^{N_d - 1} u_i^2}$, where $u / \epsilon_0$ has order of magnitude $O(1)$ matching the output of the network at initialization. The first stage residue is $e_1(x) = u(x) - u_0(x)$, and is typically a high-frequency function.

Then, the second-stage neural network is trained to approximate the normalized first stage residue $e_1 / \epsilon_1$ as a target function, where $\epsilon_1$ is the root mean square value of the error $e_1$ on the provided data points. To effectively fit the high-frequency first stage residue given the dominant frequency $f_d$ and the variance of weights in the first layer $V_{ar}$, the activation function of the first layer is taken to be the periodic, high-frequency $\sigma(x) = \cos(\kappa x)$, where its inputs are first multiplied by the optimal scale factor $\kappa = \pi f_d / \sqrt{V_{ar}}$ \cite{wang2024multi}. The re-scaling of the input to enable learning of high-frequency functions has also been implemented as adaptive activation functions \cite{jagtap2020adaptive}.

\begin{figure*}[t]
    \centering
\includegraphics[width=\linewidth]{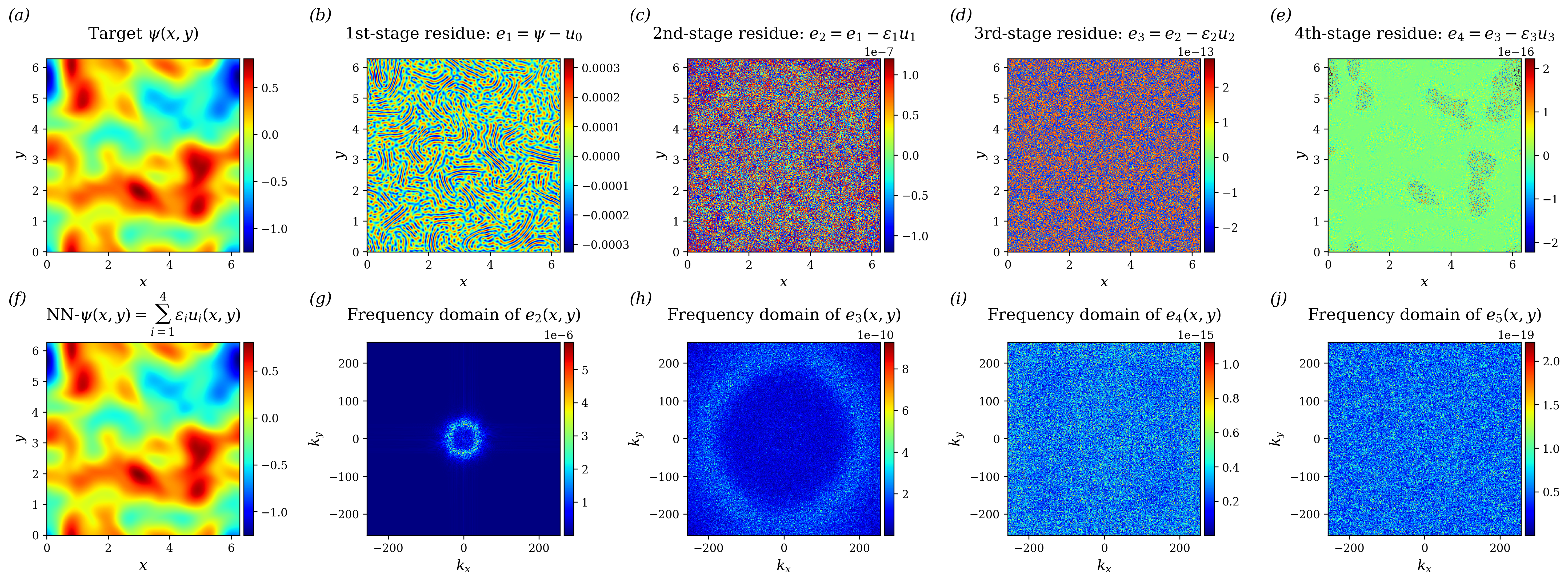}
    \caption{(a) Target function $\psi(x, y)$, a single time snapshot of the numerical solution of the stream function for the 2D incompressible Navier-Stokes equations \eqref{eq:nse1} and \eqref{eq:nse2} with Re = 2000. (b--e) The residues after each of the four stages of training, which are the target functions for the next stages. After four stages, the residue has approached machine precision $O(10^{-16})$. (g) Result of the Spectrum-Informed Multistage Neural Network (SI-MSNN) with four stages of training. (h--j) The spectral domain of each of the residues.}
    \label{fig:multistage-results}
\end{figure*}

In general, the $(n + 1)$th stage neural network $u_{n + 1}$ is the previous stage residue $e_n(x) = u(x) - \sum_{i = 1}^n \varepsilon_n u_n(x)$, normalized by its root mean square value $\epsilon_n$. The final result with $s$ stages is $\sum_{n = 1}^s \varepsilon_n u_n(x) \approx u(x).$ By tuning the neural network in each stage to capture the desired magnitudes and frequencies of the residue, the multistage neural network approach is able to reach machine precision $O(10^{-16})$ for the regression problem in one dimension. 

\subsection{$L^p$ Norm}

The typical loss function for training and evaluation in regression problems is the mean squared error (MSE), which is the squared $L^2$ norm, defined as $\norm{x}_{L_2^2} = \sum_{n = 1}^{N_d} x_n^2.$ A more general loss function uses the $L^p$ norm, where \begin{equation}\norm{x}_{L_p} = \pars*{\sum_{n = 1}^{N_d} \abs{x_n}^p}^{1/p},\end{equation} which has a comparable order of magnitude to the root mean squared error (RMSE). As $p$ increases, larger deviations are more heavily penalized. For our purposes, we use the $L^p$ norm with $p = 10$, which prevents large spikes in the residue so the next stage of the multistage neural network can learn the residue more readily. 

\subsection{Spectrum-informed initialization}

In two or more dimensions, the multistage neural network is not able to reach machine precision, and fails to capture the high frequencies present in the later stage residues. We propose a spectrum-informed initialization to replace the cosinusoidal mapping with a scale factor $\kappa$ used in \citet{wang2024multi}. Instead, here we tailor the neural network to the specific frequencies present in the dataset. Consider a dataset with input-output pairs $(\mathbf{x}_i, u_i = u(\mathbf{x}_i))$ for $\mathbf{x}_i \in [-1, 1]^2$ on a grid with $N$ points in both dimensions, where the input domain is assumed to already be transformed to $[-1, 1]$, and $u_i \in \RR$. The spectrum-informed initialization provides a targeted initialization of $B$ and $b$ for the input mapping layer $\gamma(x; B, b) = [A\cos(Bx + b)]^T$ of a neural network predicting the quantity $u(x)$ given the input coordinates $x$, where $B$ and $b$ are tunable parameters and $A$ is a fixed hyperparameter. The remainder of the neural network layers can be initialized with the usual Xavier initialization scheme, which prevents gradient vanishing or explosion \cite{glorot2010understanding}.

\begin{algorithmic}
    \begin{algorithm}
    \caption{Spectrum-informed initialization}
    
    \STATE Compute the discrete Fourier transform $\tilde u$ of $u$, where $\tilde u(\mathbf{k}) = a(\mathbf{k})e^{i\theta(\mathbf{k})}$ in polar form. Let $\alpha(\mathbf{k}) = \frac{2}{N^2}a(\mathbf{k})$ if $k_y > 0$, and $\alpha(\mathbf{k}) = \frac{1}{N^2}a(\mathbf{k})$ if $k_y = 0$ be a normalized version of the magnitude. 
    \STATE Let $\mathbf{k}^{(j)}$ be the Fourier mode corresponding to the $j$th largest magnitude $|\tilde u(\mathbf{k}^{(j)})|$, considering only the modes such that $k_y \geq 0.$ Take $\alpha^{(j)} = \alpha(\mathbf{k}^{(j)})$ and $\theta^{(j)} = \theta(\mathbf{k}^{(j)})$.
    \STATE Given a desired first layer width $n_f$, let $B = [\mathbf{k}^{(1)}, \cdots, \mathbf{k}^{(n_f)}]^T$,  $b = [\theta^{(1)}, \cdots, \theta^{(n_f)}]^T$ and $A = \left[\alpha^{(1)}, \cdots, \alpha^{(n_f)}\right]^T$
    \STATE Initialize the first layer of the neural network, which has width $n_f$, with the Fourier feature mapping $\gamma(x; B, b) = [A \cos(Bx + b)]^T$. Initialize the other parameters based on the Xavier scheme. 
    \end{algorithm}
\end{algorithmic}

The two-dimensional DFT of $u$ satisfies \begin{align} u(x, y) &= \frac{1}{N^2}\sum_{k_x, k_y} \tilde u(k_x, k_y)\exp(i(k_xx + k_yy)) \\
&= \sum_{k_x, k_y} \frac{1}{N^2}a(\mathbf{k})\exp(i(\mathbf{k}\cdot\mathbf{x} + \theta(\mathbf{k}))) \\
&= \sum_{k_x, k_y \geq 0}\alpha(\mathbf{k})\cos(\mathbf{k}\cdot\mathbf{x} + \theta(\mathbf{k})), \label{DFT of u}
\end{align}
where $\alpha({\bf k})$ are as defined in Algorithm 1. The last equality comes from the conjugate symmetry $\tilde u(\mathbf{k}) = \overline{\tilde u(-\mathbf{k})}$ of $\tilde u$, since $u$ is a real-valued function. Using the conjugate symmetry avoids embedding redundant information. 

Then, given the $n_f$ largest modes, the Fourier feature mapping, which can be considered the first layer of the neural network, is $\gamma(x) = A\cos(Bx + b) = \sum_{i \leq n_f} \alpha^{(i)}\cos(\mathbf{k}^{(i)} \cdot \mathbf{x} + \theta^{(i)})$, which contains precisely the information from the top $n_f$ modes of the Fourier transform of $u$. If $n_f = N^2 / 2$ and the neural network contains only one layer, which is the Fourier feature embedding, the single-layer neural network will exactly match the DFT representation of $u$, as in Equation \ref{DFT of u}. 

Based on the neural tangent kernel (NTK) theory \cite{jacot2018neural}, embedding these primary Fourier modes $\mathbf{k}$ allows the neural network to effectively learn the target function across the frequency spectrum. 

\section{Experiments}\label{experiments}

In this section, we experimentally demonstrate the ability of the spectrum-informed initialization to learn the target frequencies, and to train neural networks as approximators down to the limits of numerical precision. 

\subsection{Comparison with scale factor approach}

In Figure \ref{fig:msnn-comparison}, we compare the results of a three-stage neural net with the original scale factor approach and the spectrum-informed initialization, fitting a simple target function $u_g(x) = [\sin(2x + 1) - 0.5x](1 - y^2)$. For the scale factor approach, the Fourier feature embedding is of the form $\gamma(x) = \cos(\kappa w^{(0)}x + b),$ where $w^{(0)}$ are the first-layer weights following the Gaussian distribution $\mathcal{N}(0, V_{ar})$ with the variance $V_{ar}$ set by the Xavier initialization method. The scale factor $\kappa = \pi f_d / \sqrt{V_{ar}}$ is determined based on the dominant frequency $f_d$ of the target function. In contrast, the spectrum-informed initialization employs the Fourier feature embedding $\gamma(x) = A\cos(Bx + b)$ with $A, B,$ and $b$ being directly initialized by the spectral information of the target function, as described in Algorithm 1. 

Experimentally, the scale factor approach yields a 3rd-stage residue of $O(10^{-8})$ (Figure \ref{fig:msnn-comparison}$b$), whereas the spectrum-informed initialization allows the MSNN to fit the data up to $O(10^{-13})$ (Figure \ref{fig:msnn-comparison}$c$), which is five orders of magnitudes more accurate. We note that the number of iterations used in each stage of the two experiments are the same.

\begin{figure*}
    \centering
    \includegraphics[width=\linewidth]{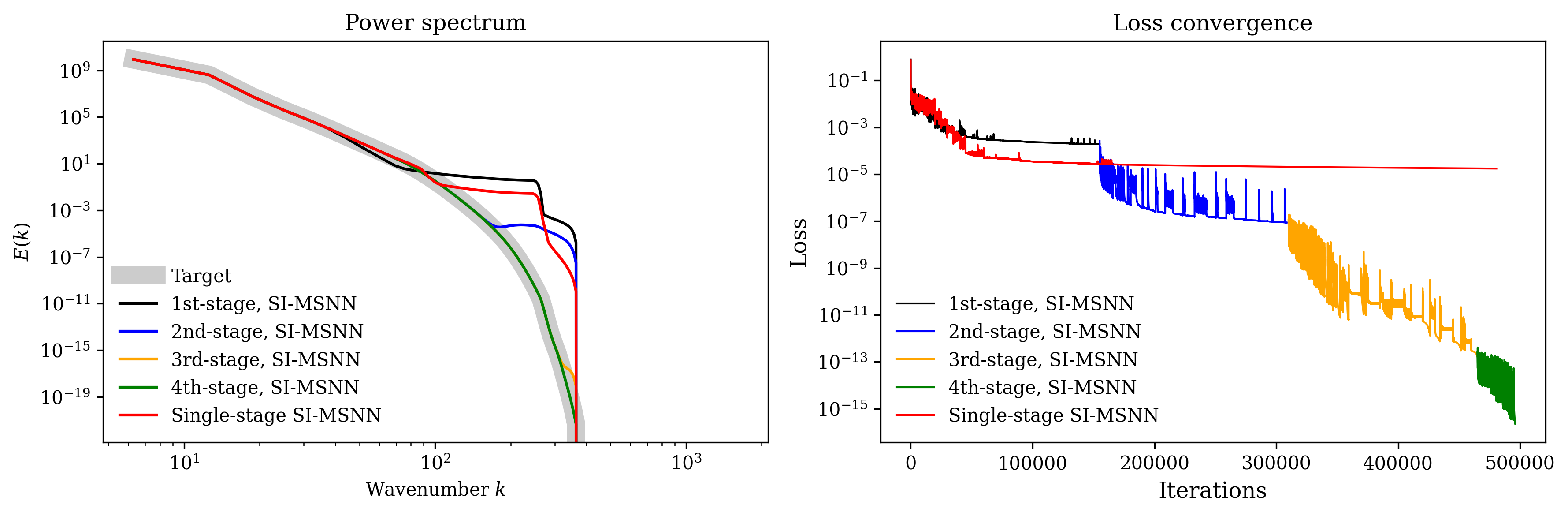}
    \caption{Left: A comparison of the power spectrum of the 2D Navier-Stokes example in Fig. \ref{fig:multistage-results} given by a single-stage SI-MSNN and a four-stage SI-MSNN, each with three hidden layers of width 30. The single-stage SI-MSNN has first layer width $n_f = 10000$. The first stage of the four-stage SI-MSNN has first layer width $n_f = 1359$, based on the number of primary Fourier modes present, and the remaining stages all have first layer width $n_f = 10000.$ Right: The loss convergence of a single-stage SI-MSNN compared to a four-stage SI-MSNN, where training terminates when machine precision is reached.}
    \label{fig:loss-convergence}
\end{figure*}

\subsection{2D incompressible Navier-Stokes} 

As an example problem, we consider fitting a snapshot of two-dimensional homogeneous isotropic decaying turbulence (2D-DHIT), which is a broadly multi-scale problem. Random Fourier features have been shown to be effective for similar multi-scale turbulence problems \cite{mojgani2024interpretable}. The Fourier coefficients of the target function decay exponentially with the wavenumber, where the high-frequency, small-scale information plays a large role in the behavior of the system. Taking the curl of the 2D incompressible Navier-Stokes equations yields the vorticity equation, which can be written in terms of the stream function $\psi$ and the vorticity $\omega$: 
\begin{equation}\frac{\partial \omega}{\partial t} + \frac{\partial \psi}{\partial y}\frac{\partial \omega}{\partial x} - \frac{\partial \psi}{\partial x}\frac{\partial \omega}{\partial y} = \frac{1}{\text{Re}}\nabla^2\omega\label{eq:nse1}\end{equation} and \begin{equation}\nabla^2\psi = -\omega,\label{eq:nse2}\end{equation}
which can model a variety of geophysical flows \cite{subel2023explaining}. Using a direct numerical simulation (DNS), on a $[0, 2\pi)^2$ grid with resolution $N = 512$ and Reynolds number $Re = 2000$, the 2D-DHIT equations are solved using a pseudo-spectral solver with third-order Runge-Kutta time stepping, with the timestep determined by the CFL condition. The initial condition is taken to be a random vorticity field with a broad banded energy spectrum \cite{mcwilliams1984emergence}. 

We perform regression with the target function as a snapshot of the stream function $\psi(\mathbf{x}, t = 1.25)$. The coordinate inputs to the neural network are re-scaled to the domain $[-1, 1]$, and the spectrum-informed initialization is based on the re-scaled coordinates. The results are transformed back to the original coordinates from $[0, 2\pi)$. In terms of experimental setup, we employ a fully-connected neural network with the first layer as the spectrum-informed Fourier feature embedding with cosine as the activation function, as well as three hidden layers with 30 units in each layer using hyperbolic tangent activation functions. The first layer contains $n_f = 1359$ units for the first-stage neural network, and $n_f = 10000$ units for the second, third, and fourth-stage neural networks, based on the number of primary Fourier modes in the target function of each stage. The spectrum-informed initialization is used for the first Fourier feature layer, and the Xavier scheme is used to initialize the other parameters, as described in Algorithm 1. The neural networks are trained with full batch gradient descent using Adam optimizer with initial learning rate of $10^{-3}$ and incorporating learning rate annealing. The models are trained for 150000 iterations, with the fourth stage terminating when machine precision is reached. 

In Figure \ref{fig:multistage-results}, the SI-MSNN method is successfully able to reduce the error down to the machine precision of a 64-bit double float, $O(10^{-16})$, after four stages of training. This result experimentally validates the ability of SI-MSNNs to approximate a target function with arbitrarily high accuracy. 

In Figure \ref{fig:loss-convergence}, the multistage setup is shown to be necessary. With a single stage of training, the loss convergence plateaus to a linear decay rate, and the neural network is not able to correct the residues relative to the target function. However, by incorporating multiple stages, with a neural network initialized based on the magnitude and spectrum of the residue, an exponential convergence rate can be maintained, allowing efficient learning. 

Moreover, the spectrum-informed initialization allows the multistage neural network to accurately represent the function across the spectral domain. Here, the two-dimensional Fourier transform of the stream function $\psi$ satisfies $\tilde \psi(k_x, k_y) = \sum_{n = 0}^{N_x}\sum_{m = 0}^{N_y} \psi(x_n, y_m)\exp(-i (k_x x_n + k_y y_m))$, where $\mathbf{k} = (k_x, k_y)$ is the wavenumber vector. Given the Fourier transform $\tilde \psi$, the angle-averaged power spectrum $E(k)$ is defined as \begin{equation}
E(k) = \sum_{k - \Delta k \leq \abs{\mathbf{k}} \leq k + \Delta k} \abs{\tilde \psi(\mathbf{k})}^2,
\end{equation}
for a given spectral band $\Delta k$ \cite{kag2022physics}. The spectral band is defined so that the wavenumber $k$ falls into 200 bins.

In Figure \ref{fig:loss-convergence}, as the number of stages increases, the SI-MSNN is able to match the power spectrum $E(k)$ of the target function $\psi(x, y)$ up to increasing wavenumbers. Successfully fitting both the target function and its power spectrum is important for multi-scale scientific problems, and the capacity to do so is shown with this 2D Navier-Stokes example. 

\section{Discussions}\label{discussions}

We introduce a novel spectrum-informed initialization, allowing efficient training of neural networks for solving the regression problem to machine precision. By utilizing the spectral biases of neural networks, a spectrum-informed Fourier embedding of the input allows the neural network to learn in the spectral domain and converge rapidly. The spectrum-informed initialization for multistage neural networks (SI-MSNN) allows the neural network to fit target functions down to machine precision, even for multi-scale target functions, as validated experimentally. The results using the SI-MSNN demonstrate that this approach can achieve residues many orders of magnitude smaller than state-of-the-art approaches, down to machine precision. 

In the future, we plan on applying the spectrum-informed initialization for scientific machine learning problems requiring precision as high as possible. In particular, we propose using the spectrum-informed initialization for multi-stage physics-informed neural networks as partial differential equation solvers \cite{wang2024multi,raissi2019physics}, where the boundary and initial conditions are the error-free data provided, along with a set of governing equations. Our work provides a promising approach towards precision machine learning. 

\bibliography{paper}
\bibliographystyle{icml2024}

%%%%%%%%%%%%%%%%%%%%%%%%%%%%%%%%%%%%%%%%%%%%%%%%%%%%%%%%%%%%%%%%%%%%%%%%%%%%%%%
%%%%%%%%%%%%%%%%%%%%%%%%%%%%%%%%%%%%%%%%%%%%%%%%%%%%%%%%%%%%%%%%%%%%%%%%%%%%%%%
% APPENDIX
%%%%%%%%%%%%%%%%%%%%%%%%%%%%%%%%%%%%%%%%%%%%%%%%%%%%%%%%%%%%%%%%%%%%%%%%%%%%%%%
%%%%%%%%%%%%%%%%%%%%%%%%%%%%%%%%%%%%%%%%%%%%%%%%%%%%%%%%%%%%%%%%%%%%%%%%%%%%%%%
\newpage
\appendix
\onecolumn
% \section{Appendix A}

% You can have as much text here as you want. The main body must be at most $8$ pages long.
% For the final version, one more page can be added.
% If you want, you can use an appendix like this one.  

% The $\mathtt{\backslash onecolumn}$ command above can be kept in place if you prefer a one-column appendix, or can be removed if you prefer a two-column appendix.  Apart from this possible change, the style (font size, spacing, margins, page numbering, etc.) should be kept the same as the main body.
%%%%%%%%%%%%%%%%%%%%%%%%%%%%%%%%%%%%%%%%%%%%%%%%%%%%%%%%%%%%%%%%%%%%%%%%%%%%%%%
%%%%%%%%%%%%%%%%%%%%%%%%%%%%%%%%%%%%%%%%%%%%%%%%%%%%%%%%%%%%%%%%%%%%%%%%%%%%%%%

\end{document}